\pdfoutput=1
\documentclass[11pt]{article}

\usepackage{EMNLP2022}

\usepackage{times}
\usepackage{latexsym}
\usepackage{graphicx}
\usepackage{booktabs}
\usepackage{multirow}
\usepackage{verbatim}
\usepackage{amsmath}
\usepackage{bbold}
\usepackage{color}
\usepackage{siunitx}
\usepackage{colortbl}
\usepackage{makecell}
\usepackage[ruled,linesnumbered]{algorithm2e}
\usepackage{algpseudocode} 
\usepackage{subcaption}
\usepackage{bbm}
\definecolor{maroon}{cmyk}{0,0.87,0.68,0.32}

\usepackage[T1]{fontenc}
\usepackage[utf8]{inputenc}
\usepackage{microtype}

\usepackage{inconsolata}

\title{FRSUM: Towards Faithful Abstractive Summarization via Enhancing Factual Robustness}

\author{ \parbox{\linewidth}{\centering Wenhao Wu\textsuperscript{1}\thanks{\ \ Work is done during an internship at Baidu Inc.}, Wei Li\textsuperscript{2}, Jiachen Liu\textsuperscript{2}, Xinyan Xiao\textsuperscript{2}, Ziqiang Cao\textsuperscript{3}, Sujian Li\textsuperscript{1}\thanks{\ \ Corresponding author.}, Hua Wu\textsuperscript{2} }\\
    \textsuperscript{1}Key Laboratory of Computational Linguistics, MOE, Peking University \\
  \textsuperscript{2}Baidu Inc., Beijing, China \\
  \textsuperscript{3}Institute of Artificial Intelligence, Soochow University, China\\
  \texttt{\{waynewu,lisujian\}@pku.edu.cn}, \texttt{\{zqcao\}@suda.edu.cn}\\
  \texttt{\{liwei85,xiaoxinyan,liujiachen,wu\_hua\}@baidu.com}\\
 }

\begin{document}
\maketitle
\begin{abstract}
Despite being able to generate fluent and grammatical text, current Seq2Seq summarization models still suffering from the unfaithful generation problem.
In this paper, we study the faithfulness of existing systems from a new perspective of factual robustness which is the ability to correctly generate  factual information over  adversarial unfaithful information.
We first measure a model's
factual robustness by its success rate to defend against adversarial attacks when generating factual information.
The factual robustness analysis on a wide range of current systems shows its good consistency with human judgments on faithfulness.
Inspired by these findings, we propose to improve the faithfulness of a model by enhancing its factual robustness.
Specifically, we propose a novel training strategy, namely FRSUM, which teaches the model to defend against both explicit adversarial samples and implicit factual adversarial perturbations.
Extensive automatic and human evaluation results show that FRSUM consistently improves the faithfulness of various Seq2Seq models, such as T5, BART.
\end{abstract}
%We first define the measurement of a model's factual robustness as its success rate to defend against adversarial attacks when generating factual information. 
\section{Introduction}
Abstractive summarization aims to produce fluent, informative, and faithful summaries for a given document. 
Benefiting from large-scale pre-training techniques, recent abstractive summarization systems are able to generate fluent and coherent summaries \cite{DBLP:conf/nips/00040WWLWGZH19,lewis-etal-2020-bart,DBLP:conf/ijcai/XiaoZLST0W20, DBLP:conf/icml/ZhangZSL20}.
However, challenges remain for this task.
One of the most urgent requirements is to improve ``faithfulness" or ``factual consistency" of a model, which requires the generated text to be not only human-like but also faithful to the given document \cite{maynez-etal-2020-faithfulness}.
An earlier study observes nearly 30\% of summaries suffer from this problem on the Gigawords dataset \cite{DBLP:conf/aaai/CaoWLL18}, while recent large-scale human evaluation concludes that 60\% of summaries by several popular models contain at least one factual error on XSum    \cite{pagnoni-etal-2021-understanding}.  
These findings push the importance of improving faithfulness of summarization to the forefront of research.

Many recent studies focus on improving the faithfulness of summarization models, which can be mainly divided into three types.
The first  type
modifies the model architecture to introduce pre-extracted guidance information as additional input \cite{DBLP:conf/aaai/CaoWLL18,dou-etal-2021-gsum,zhu-etal-2021-enhancing}, while the second type relies on a post-editing module to correct the generated summaries \cite{dong-etal-2020-multi-fact,DBLP:conf/naacl/ChenZSR21}.
The last type takes advantages of auxiliary tasks like entailment \cite{DBLP:conf/coling/LiZZZ18}, and QA  \cite{DBLP:conf/acl/HuangWW20,DBLP:conf/acl/NanSZNMNZWAX20} on faithfulness.
Different from previous studies, this work focuses on refining the training strategy of Seq2Seq models to improve %their 
faithfulness universally without involving any extra parameters, post-editing procedures and external auxiliary tasks.

In this paper, we study the faithfulness problem of Seq2Seq models from a new perspective of factual robustness, which is the robustness of generating factual information.
We first define factual robustness as the model's ability to correctly generate  factual information over  adversarial unfaithful information.
%By measuring the factual robustness of a wide rage of current summarization systems, we demonstrate 
Following this definition, we analyze the factual robustness of a wide range of Seq2Seq models by measuring their success rate to defend against adversarial attacks when generating factual information.
The analysis results (see Table \ref{tab:adv}) demonstrate good consistency between models' factual robustness and their faithfulness according to human judgments, and also reveal that current models are vulnerable to generating different types of unfaithful information.
For example, the robustness of generating numbers in the XSum dataset for most Seq2Seq models is very weak.
Inspired by the findings above, we propose a novel faithful improvement training strategy, namely FRSUM, which improves a model's faithfulness by enhancing its factual robustness.
Concretely, FRSUM teaches the model to defend against adversarial attacks by a novel factual adversarial loss, which constrains the model to generate correct information over the unfaithful adversarial samples.
%As a result, the model is aware of generating the correct information over the unfaithful ones.
To further improve the generalization of FRSUM, we add factual adversarial perturbation to the training process which induces the model to generate unfaithful information.
In this way, FRSUM not only requires the model to defend against explicit adversarial samples but also become insensitive to implicit adversarial perturbations.
Thus, the model becomes more robust in generating factual spans, and  generates fewer errors during inference.
Moreover, FRSUM is adaptive to all Seq2Seq models.

Extensive experiments on several state-of-the-art Seq2Seq models demonstrate the effectiveness of FRSUM, which  improves the faithfulness of various Seq2Seq models while maintaining their informativeness.
Besides automatic evaluation, we also conduct fine-grained human evaluation to analyze different types of factual errors.
The human evaluation results also show that FRSUM greatly reduces different types of factual errors. %specific type of factual errors. 
Especially, when applying to T5, our method reduces 17.5\% and 49.1\% of target factual errors on the XSum and CNN/DM datasets, respectively.
Our contributions can be summarized as the following three points.
\begin{itemize}
\item We study the problem of unfaithful generation from a new perspective, factual robustness of Seq2Seq models, which is found consistent with faithfulness of summaries. %by human judgments.
\item We propose a new training method, FRSUM, which improves the factual robustness and faithfulness of a model by defending against both explicit and implicit adversarial attacks.
\item Extensive automatic and human evaluations validate the effectiveness of FRSUM and also show that FRSUM greatly reduces different types of factual errors.
%demonstrate that it's promising to improve and assess faithfulness of Seq2Seq models by designing richer adversarial samples.
\end{itemize}

\section{Related Work}

\subsection{Faithfulness of  Summarization}
Studies of faithfulness mainly focus on how to improve the faithfulness of an abstractive summarization model~\cite{https://doi.org/10.48550/arxiv.2203.05227}.
Though it is  challenging, some recent works propose various methods to study this problem, which can be summarized as follows.
Some  typical methods use pre-extracted information from input the document as additional input \cite{dou-etal-2021-gsum}, like triplets \cite{DBLP:conf/aaai/CaoWLL18}, keywords \cite{DBLP:journals/corr/abs-2012-04281}, knowledge graphs \cite{DBLP:conf/acl/HuangWW20,zhu-etal-2021-enhancing,wu-etal-2021-bass} or extractive summaries \cite{dou-etal-2021-gsum}.
These methods encourage the model to copy from the faithful guidance information.
Another type of popular method focuses on designing a post-editing module, like a QA  model  \cite{dong-etal-2020-multi-fact}, or a BART-based selection model  \cite{DBLP:conf/naacl/ChenZSR21}.
But these methods are much less time efficient during inference, thus hard to be used in real-world applications.
Some other works apply Reinforcement Learning (RL) based methods, especially policy gradient, which utilize a variety  of factual-relevant tasks for calculating rewards, such as information extraction \cite{DBLP:conf/acl/ZhangMTML20}, entailment \cite{DBLP:conf/coling/LiZZZ18}, QA \cite{DBLP:conf/acl/HuangWW20,DBLP:conf/acl/NanSZNMNZWAX20}. 
This type of methods suffer from the  high-variance training of RL.
\begin{figure*}
\centering
\includegraphics[scale=0.22]{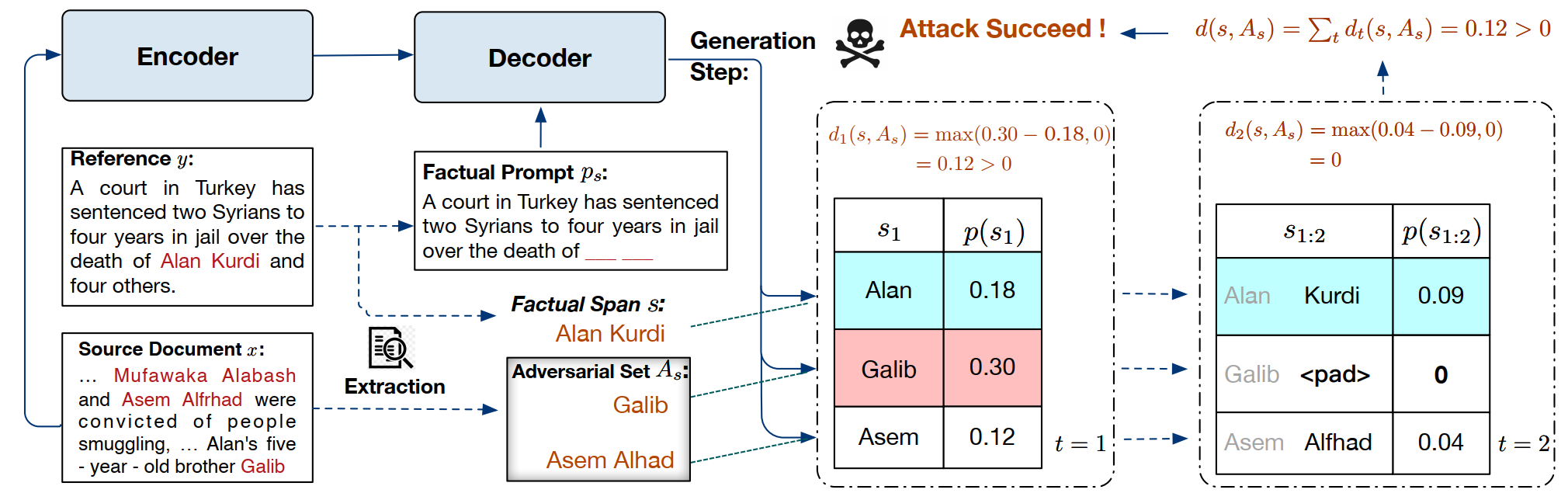}
\caption{
Procedure of an  adversarial attack on a two-token entity span. 
After extracting a factual span $s$, a factual prompt $p_s$, and a set of corresponding adversarial  samples  $A_s$, we calculate the  probability of  generating $s$ and spans in $A_s$ given $p_s$.
Based on the probability,  we check whether  this attack succeeds, according to Equation ~\ref{const_dist}.}
\label{fig:attack}
\end{figure*}

\subsection{Adversarial Attacks for Text}
Though deep neural networks (DNNs) have shown significant performance on various tasks, a series of studies have found that adversarial samples by adding imperceptible perturbations could easily fool DNNs \cite{DBLP:journals/corr/SzegedyZSBEGF13,
DBLP:journals/corr/GoodfellowSS14}.
These findings not only reveal potential security threats to DNN-based systems, but also show that training with   adversarial attacks can enhance the robustness of a system \cite{DBLP:conf/sp/Carlini017}.
Recently, a large amount of studies focus on adversarial attacks for a variety of NLP tasks, such as text classification \cite{DBLP:conf/acl/EbrahimiRLD18,DBLP:conf/naacl/GilCGB19}, question answering \cite{DBLP:conf/emnlp/JiaL17,DBLP:conf/acl/GanN19} and natural language inference \cite{DBLP:conf/conll/Minervini018,DBLP:conf/emnlp/LiMGXQ20}. 
Because of the discrete nature of language, these works mainly apply the methods of inserting, removing, or deleting different levels  of text units (char, token, sentence) to build adversarial samples \cite{ren-etal-2019-generating,DBLP:conf/acl/ZangQYLZLS20}.  
Besides the aforementioned language understating tasks, some recent works also apply adversarial attacks on language generation.
\citet{DBLP:conf/acl/ChengJM19} applies adversarial attacks on both encoder and decoder to improve the performance of translation.
Seq2Sick focuses on designing adversarial samples to attack SeqSeq models for evaluating their robustness on informativeness \cite{DBLP:conf/aaai/ChengYCZH20}. 
Compared with previous works, we are the first to study the problem of unfaithful generation from the perspective of robustness.
\section{Factual Robustness on Seq2Seqs}\label{fr}
\begin{table*}[]
    \centering
    \begin{tabular}{l|SSScS|SSSSS}
   \Xhline{2\arrayrulewidth}

    \multirow{2}{*}{System} &
      \multicolumn{5}{c|}{XSum}  &
      \multicolumn{5}{c}{CNN/DM} \\
     
     &\small{Mix\%}$\downarrow$ &\small{Ent\%}$\downarrow$ & \small{Num\%}$\downarrow$ &\small{R-L}$\uparrow$&\small{Incor\%}$\downarrow$ &\small{Mix\%}$\downarrow$&\small{Ent\%}$\downarrow$ & \small{Num\%}$\downarrow$ &\small{R-L}$\uparrow$&\small{Incor\%}$\downarrow$\\
   \Xhline{2\arrayrulewidth}
   TransS2S&53.1&54.0&52.1&24.0&96.9&48.0&50.8&40.5&35.9&74.8\\
   BERTSum&40.1&36.0&47.2&31.2&83.7&33.4&36.2&29.5&38.8&27.2\\
   T5&37.3&33.2&43.4&33.1&82.0&37.5&40.2&31.9&40.2&26.7\\
   BART&26.7&25.0&31.6&36.9&66.7&29.0&32.2&23.8&40.5&24.7\\
   PEGASUS&22.4&20.0&29.0&39.1&60.7&28.3&29.6&22.2&40.5&13.3\\
   \Xhline{2\arrayrulewidth}

    \end{tabular}
    \caption{The factual robustness of different systems on CNN/DM and XSum datasets. Ent\% and Num\% are the success rate $E$ of adversarial attack on entity and number spans, respectively.
    Mix\% is the average success rate $E$ of attacking both the number and entity spans. 
    R-L is the abbreviation of ROUGE-L listed aside for reference.
    Incor\% is incorrect ratio of generated summaries annotated by humans.
    The Pearson Correlation Coefficient and Spearman Correlation Coefficient between Mix$\%$ and Inroc$\%$ are 0.57 and 0.60, respectively.
    }
    \label{tab:adv}
\end{table*}
In this section, we introduce the definition and  measurement of factual robustness.
The factual robustness is defined as   the ability of a Seq2Seq model to correctly generate    factual  information  over    adversarial unfaithful information.  
Formally, given a document $x$, a faithful summary $y$ and a set of unfaithful adversarial samples $A_s$, a model with high factual robustness should satisfy:
\begin{equation}
    p(y|x) > \max_{s^a \in A_s} p(s^a|x)
\end{equation}
where $p(y|x)$ is the probability of generating $y$ given $x$, $s^a$ is the adversarial sample in $A_s$.
We adopt a process similar to  adversarial attacks to measure the factual robustness of a Seq2Seq model.
Extending from the conventional  adversarial attack framework, we take the generation process of a  factual information span as the target for attack.
After constructing a set of adversarial samples, we check whether an attack succeeds by comparing the generation probabilities between the  span and adversaries.
We then define the measurement of factual robustness as the success rate of a model to defend against these attacks in a corpus.
Following this definition, we measure the factual robustness of current models and analyze their relations with faithfulness. 
%We then  measure factual robustness as the success rate of a model to defend against these attacks in a corpus. Following this measure, we analyze the relations between the factual robustness and faithfulness for current models. 
%\In the following, we introduce the formal definition and measurement of factual robustness .
\subsection{Measurement of Factual Robustness}
In this section, we measure the factual robustness of Seq2Seqs by adversarial attacks.
Though adversarial attacks have been well-studied in classification tasks, 
it is still not straightforward to directly adapt them to  text generation models.
%it is not straightforward to be directly to text generation models.
Different from attacking on a single label prediction in classification tasks, we consider the multi-step token predictions when  generating a span of information.
%%% we try to apply attack on multi-step token predictions when generating a span of information.
%%%

\paragraph{Factual Span and Factual Prompt}  Given a document $x$ and its reference  with $m$ tokes   $y=\{y_1,y_2,\dots,y_m\}$, we define a factual span $s$ as the elementary unit of factual information, which can represent various types of facts. %and is utilized as the target for factual adversarial attack. 
%We define a \textit{factual span} $s$ as a span of tokens that represents a piece of specific factual information, which can represent various types of facts.
As the first study on factual robustness, we only analyze entity and number spans which are the most common types of information errors in existing summarization models.
%Following previous studies \cite{dong-etal-2020-multi-fact,DBLP:conf/emnlp/KryscinskiMXS20} , we only analyze entity and number spans.
After extracting a span $s$, we define the prefix before $s$ in  $y$ as \textit{factual prompt} $p_s$, based on which the model should generate span $s$ correctly.
\paragraph{Adversarial Sample} $s^{a}$ is a span that makes the information $[p_s,s^{a}]$ contradict the input $x$.
It is used to attack the generation process of $s$.
Previous study finds that intrinsic  hallucinations are the most frequent factual errors in Seq2Seq models \cite{maynez-etal-2020-faithfulness}.
This kind of factual errors usually occur when the model confuses other information presented in the input document with the target information during generation.
Thus, in this study, we construct \textit{a set of adversarial samples} $A_s$ by extracting entity and number spans from the source document $x$ that are different from  the target span s: $A_s=\{s^a|s^a\in x \&  s^a\neq s\}$ to introduce intrinsic hallucinations.
\paragraph{Adversarial Attack}  We measure factual robustness by an adversarial attack process which utilizes the above adversarial samples.
%We introduce the the adversarial attack process we apply to measure factual robustness.
Specifically, given the input $x$ and a factual prompt $p_s$, 
 we apply adversarial attacks to auto-regressively generate $s$ by using the adversarial samples in $A_s$. 
%we apply adversarial attacks on the process of auto-regressively generating $s$ by using the adversarial samples in $A_s$. 
In every generation step, we check whether the model has the highest probability to generate the prefixes of $s$.
Following conditional probability, in step $t$, the probability of generating the first $t$-token prefix of $s$ ($t\leq|s|$, $|s|$ denotes the length of $s$) is:
\begin{equation} \label{span_gen}
    p(s_{1:t}|p_s,x,\theta)= \prod_{i=1}^{t}p(s_i|s_{1:i-1},p_s,x,\theta)
\end{equation}
where $s_i$ and $s_{1:i}$ are respectively the $i$-th token and first $i$-tokens of $s$, and  $\theta$ denotes the model parameters.
In the following, Eq.~\ref{span_gen} is abbreviated as $p(s_{1:t}) $.
Based on it, % the definition of $p$, 
we compare the probability of generating  the target factual span $s_{1:t}$ and adversarial samples $s^a_{1:t}$ as:
\begin{equation}\label{pairwise}
    d_t(s,A_s)=\max_{s^a\in A_s}(\max(p(s^a_{1:t}) -p(s_{1:t}),0))
\end{equation}
which measures the tendency of generating unfaithful spans in adversarial samples over the factual span.
To measure the full generation process, we average %the probability contrast 
$d_t(s,A_s) $ of the total $|s|$ generation steps:
\begin{equation}\label{const_dist}
    d(s,A_s) =  \frac{1}{|s|}\sum_{t=1}^{|s|} d_t(s,A_s)
\end{equation}
For the  adversarial samples with different length $|s^a|\neq|s|$, $s^a$ is truncated or padded to  $|s|$, where the  probability of generating the pad token is  0.
In this way, each step we can compare the probability of generating prefixes of spans with the same length.
%In any step, 
If any adversarial sample has a higher probability, then $d(s,A_s)>0$, indicating the success of this adversarial attack.
An example of a successful  adversarial attack is illustrated in Figure \ref{fig:attack}.
In the first step, the model has a higher probability of generating the token ``\textit{
Galib}'' in adversarial samples instead of  ``\textit{Alan}'', so the adversarial attack succeeds.
\paragraph{Factual Robustness} \quad Following the definition above, we measure the factual robustness of a model via its success rate of attacks in a corpus.
Given a test set $D$ and a model with parameters $\theta$, following Equation~\ref{const_dist}, the success rate of adversarial attack on $D$ is calculated as:
\begin{equation}\label{factual_robust}
    E =\frac{\sum_{x,y \in D}\sum_{s \in y}\mathbb{1}[d(s,A_s)>0]}{\sum_{y \in D}C_s(y)} 
\end{equation}
where $C_s(y)$ is the number of factual spans in the reference $y$, and $\mathbb{1}$ is the indicator function. 
Obviously, lower $E$ indicates better factual robustness.

\subsection{Factual Robustness and Faithfulness}
Following  Eq.\ref{factual_robust}, we  measure factual robustness of current SOTA Seq2Seq summarization systems and analyse their relations with faithfulness.
We report  both factual robustness and  faithfulness of models  on different datasets in Table \ref{tab:adv}.
Details about these models and datasets are introduced in \S \ref{exp:info_cnn}.
We evaluate the factual robustness for two  kinds of factual spans, i.e. entity and number.
Their corresponding success rates of adversarial attacks are denoted as Ent$\%$ and Num$\%$.
Mix$\%$ is the average success rate of attacking all the entity and number spans in the reference summary $y$.
Incor\% denotes the ratio of unfaithful summaries judged by humans\footnote{Incor\% annotation of T5 comes from $\S$~\ref{exp:info_cnn}, while TransS2S and BERTSum come from \citet{pagnoni-etal-2021-understanding}, BART and PEGASUS come from \citet{DBLP:conf/emnlp/Cao021}.}.

From  Mix\% and Incor\% reported in Table \ref{tab:adv}, we can conclude that factual robustness and faithfulness have good consistency: the more factually robust the model is (lower Mix\%) the better faithfulness the generated summaries exhibit (lower Incor\%).
Specifically, the Pearson Correlation Coefficient and Spearman Correlation Coefficient between factual robustness (Mix$\%$)  and faithfulness (Incor\%) are 0.57 and 0.60, respectively, which also show the great potential of utilizing factual robustness for faithfulness assessment.
We also draw several other conclusions based on the results. 
Firstly, considering the simplicity of our adversarial samples, current systems are still vulnerable in factual robustness.
Even the current SOTA models PEGASUS and BART fail to defend nearly 30\% of the attacks.
It can be further supposed that these models will have a lower factual robustness when defending against stronger adversarial samples.
Secondly, a better pre-training strategy not only largely improves ROUGE scores but also improves the factual robustness and faithfulness, which is also confirmed by human evaluations \cite{maynez-etal-2020-faithfulness}.
Lastly, different types of factual spans perform differently regarding factual robustness.
Generating numbers is more challenging in XSum than CNN/DM because it requires the model to  comprehend and rewrite the numbers in the summaries rather than just copying spans contained in the input.

\section{FRSUM}
In the previous section, we introduce factual robustness  and reveal its relation with faithfulness.
We also discover that current systems are not  robust enough in generating factual spans.
Based on these findings, it is natural to improve a model's faithfulness by enhancing its factual robustness.
Thus, we propose FRSUM, which is a  training strategy to improve the faithfulness  of Seq2Seq models  by enhancing   their factual robustness.
FRSUM is composed of a factual adversarial loss and factual adversarial perturbation.
Factual adversarial loss encourages the model   to defend against explicit adversarial samples. 
Factual adversarial perturbations further apply implicit  factual-relevant adversarial permutations to the previous procedure to enhance the factual robustness.
We follow the notations in \S \ref{fr} to introduce FRSUM in detail.

FRSUM can be applied to all kinds of Seq2Seq models which are composed of an encoder and a decoder.
Following the common Seq2Seq architecture, we apply Negative Log Likelihood  (NLL) in the training process to generate fluent summaries.
Given a document $x$ and its reference $y$, the encoder first encodes input document $x=(x_1,x_2,\dots,x_n)$ of length $n$ into hidden representations $h$.
After that, the decoder computes the NLL based on $h$ and $y$:
\begin{equation}\label{mle_loss}
    \mathcal{L}_{nll}(\theta) =-\frac{1}{m} \sum_{t=1}^{m} \log p(y_t|y_{<t},h,\theta)
\end{equation}

\subsection{Factual Adversarial Loss}
In addition to NLL, we further propose the factual adversarial loss to enhance   the model's factual robustness.
As introduced in \S \ref{fr}, we apply the success rate $E$ of adversarial attack   to measure a model's factual robustness.
Similarly, we can also enhance factual robustness by optimizing  $E$.
Because  Eq.~\ref{factual_robust} is discrete and intractable for gradient-based optimization, we apply the probability contrast between $s$ and $A_s$ (as in Eq.~\ref{pairwise}) for optimization instead. %$d(f,E_f^-)$ between s and $E_f^-$ for learning.
We first modify the probability contrast between two samples $s$ and $s^a$ by further adding a constant margin $\gamma>0$ to adjust the degree of contrast:
\begin{equation}
    d_t(s^a,s,\gamma) = \max(lp(s^a_{1:t}) - lp(s_{1:t}) +\gamma,0) \nonumber
\end{equation}
where $lp$ denotes the logarithm of  the original $p$, $t$ denotes the $t$-th generation step,  consistent with previous sections.
In this way, we encourage the model to generate faithful content over the adversaries by a margin in probability. 
Then, we expand the above pairwise probability contrast to a set of adversarial samples $A_s$ and further compute the factual adversarial loss:
\begin{equation}\label{fc_loss}
    \mathcal{L}_{fa} = \frac{1}{C_s(y)} \sum_{s \in y}\frac{1}{|s|}\sum_{t=1}^{|s|} \max_{s^a\in A_s} d_t(s^a,s,\gamma)
\end{equation}

\subsection{Factual Adversarial Perturbation}
%In the factual adversarial loss,  we encourage the generation of spans in the original $y$ which is also optimized in the  NLL loss.
%Thus, we add perturbations to the factual adversarial loss to separate it from  NLL and also make it more challenging to train. 
Besides defending against explicit adversarial samples, we further apply implicit adversarial perturbations to enhance generalization of factual robustness~\cite{DBLP:conf/iclr/MadryMSTV18}.
We propose factual adversarial perturbations and add them to the training process,  which induce the model to have a higher probability to generate unfaithful information. 
In this way, FRSUM not only requires the model to defend against explicit adversarial samples but also become insensitive to implicit adversarial perturbations.  
%In this way, optimizing factual adversarial loss acquires the model to defend against adversarial attack under the obstruction by perturbations.
%Before training the model to defend against adversarial samples, we add  perturbations to this process to induce the model to generate unfaithful content.
%$In this way, the training process is more challenging and further improves the factual robustness.   
Formally, the purpose of the perturbation is to disturb the generation of factual span $s$ as much as possible.
We measure the quality of generating  s by its NLL  given the factual prefix $p_s$:
\begin{equation}
     l_s(\theta,h) = -\sum_{t=1}^{|s|} \log p(s_t|s_{1:t-1},p_s,h, \theta)
\end{equation}
%where $f_t$ is the $t$-th token of s.
For the simplicity of implementation, we add perturbation $\delta=[\delta_1\dots,\delta_n]$  on the encoded hidden states $h$.
Following the definition of adversarial perturbation, the expected perturbation  should satisfy the following condition:
\begin{equation}
    \delta =  \mathop{\arg\max}\limits_{\delta',||\delta'||\leq\epsilon} l_s(\theta,h+\delta')
\end{equation}
where the norm of $\delta'$ is constrained to be smaller than  $\epsilon$.
%where $\epsilon$ constrains $\delta'$.
We follow \citet{DBLP:journals/corr/GoodfellowSS14} to approximate $\delta$  by the first-order derivative of $l_s$, because the exact solution for $\delta$ is intractable in deep neural networks:
\begin{gather}
    \delta = \nabla_h l_s(\theta,h) / \|\nabla_h l_s(\theta,h)\| \\
    \widehat h = h+\tau*\delta \label{permu}
\end{gather}
\noindent where $\widehat h$ is the hidden representation after perturbation, and $\tau$ is the update step.
We then replace $h$ with $\widehat h$  to predict the probability of generating $s$ and $s^a$ to compute $\mathcal{L}^p_{fa}$  following  Eq.\ref{fc_loss}, which is the perturbed version of $\mathcal{L}_{fa}$,    
%After getting the perturbated hidden state, we use it in the following steps by replacing $\widehat h$ with $h$.

\subsection{Training Procedure}
The overall loss function of FRSUM is:
\begin{equation}
     \mathcal{L} =  \mathcal{L}_{nll} +  \eta *\mathcal{L}^p_{fa} 
\end{equation}
where $\eta \in [0,1]$ balances the NLL and factual adversarial loss.
We gradually increase  the difficulties of training by slowly increasing $\tau$ in Equation~\ref{permu}:
\begin{equation}
    \tau = \min(\max((epoch-S),0)*0.1,0.5)
\end{equation}
where $epoch$ is the number of current training epoch and $S$ is the initial epoch to apply explicit adversarial perturbations.
When $epoch > S$,  $\tau$  gradually increase till the maximum of 0.5.

The whole training process is illustrated in Algorithm 1.
For a given document-reference pair $(x,y)$, we first extract and sample an entity or numeric span $s$ from $y$  and its corresponding adversarial set $A_s$ from $x$ (line 2-3), where $Sample(a,b)$ indicates sampling $b$ samples from set $a$, $E()$ indicates the extraction of entity or number.
After the model calculated $\mathcal{L}_{nll}$ (line 5-7), we add adversarial perturbations to $h$ (line 9-10), where $s_s$ and $s_e$ are the start position and end position of $s$ in $y$. 
After that,  we apply $\hat{h}$ to calculate factual contrast loss $\mathcal{L}^p_{fc}$ based on the perturbated hidden state $\hat{h}$ (line 12-16).
Finally, we use the final output loss $\mathcal{L}$ for training.
\begin{algorithm}
  \small
  \SetKwInOut{Input}{Input}\SetKwInOut{Output}{Output}

  \Input{Document $x$, Reference $y$, Entity and Number extractor E().}
  \Output{Training loss $\mathcal{L}$}
   \Comment{Data Pre-processing}\\
  $s,p_s \leftarrow  Sample(E(y),1) $\;
 $A_s \leftarrow Sample(E(x)\setminus s$, 10)\\
  \Comment{NLL Loss}\\
 $h=Encoder(x)$
  \BlankLine
$P_{tgt}=Decoder(h,y)= [p_1,p_2,\dots,p_m]$\;
 \BlankLine
$\mathcal{L}_{nll} = -\frac{1}{m}\sum_{i=1}^m\log P_{tgt}[i]$\;
  \Comment{Factual Relevant Permutation}\\
$l_s(\theta,h) =-\frac{1}{|s|}\sum_{i=s_{s}}^{s_{e}}\log P_{tgt}[i]$
\BlankLine
$\hat{h} =  h+\epsilon *\nabla_h l_s/|\nabla_h l_s| $\\
  \Comment{Factual Contrast Loss}\\
$p(f) = Decoder(\hat{h},[p, f])$
\BlankLine

\For {$s^a$ in $A_s$}{
$p(s^a) = Decoder(\hat{h},[p_s,s^a])$
}

\BlankLine
$\mathcal{L}^p_{fc} \leftarrow$  Eq.\ref{fc_loss} with $p(s)$,$\{p(s^a)|a^a\in A_s\}$\\
 \Comment{Output Loss}\\
$\mathcal{L} = \mathcal{L}_{nll}+ \eta* \mathcal{L}^p_{fc}$
  \caption{FRSUM}\label{algo_disjdecomp}
\end{algorithm}\DecMargin{1em}
\begin{table}[t]
    \small
   \setlength{\tabcolsep}{1.3mm}
   \renewcommand{\arraystretch}{1.2}%调行距
  \begin{tabular}{l|ccccc}
   \Xhline{2\arrayrulewidth}
     &\textbf{CC}&\textbf{SC}& R-1&R-2&R-L  \\
      \Xhline{2\arrayrulewidth}
       \rowcolor[gray]{.8}
      TransS2S&83.27&79.10& 39.30&17.27&35.89\\
      Split Encoders&73.11& -&38.83&16.51&35.71\\
      Fact Correction&82.82&-& 39.87&17.50&36.80\\
      FRSUM&$\textbf{83.75}^\dagger$&$\textbf{79.97}^\dagger$&$\textbf{41.25}$&$\textbf{18.96}$&$\textbf{37.99}$\\
      \Xhline{2\arrayrulewidth}
      \rowcolor[gray]{.8}
       BERTSUM&75.73&71.66&41.72&19.39&38.76\\
       Split Encoders&76.43&-&39.78&17.87&37.01\\
       Fact Correction&78.69&-&41.13&18.58&38.04\\
       FRSUM&$\textbf{77.18}^\dagger$&\textbf{72.21}&41.59&19.03&38.66\\
      \Xhline{2\arrayrulewidth}
%    \rowcolor[gray]{.8}
%    PEAGASUS&70.00&69.95& 44.27&21.47&41.05\\
%    ContrastSel&74.03&30.43&43.41&20.27&40.30\\
%    CLIFF&76.08&75.04&30.43&19.75&39.17\\
%    FRSUM&&&\\
%     \Xhline{2\arrayrulewidth}
    \rowcolor[gray]{.8}
  BART&80.66&78.66&\textbf{43.88}&\textbf{20.93}&\textbf{40.57}\\
  ContrastSel&83.23 &74.03&42.66&19.82&39.34\\
  CLIFF&78.13&77.59&43.92&20.95&40.60\\
  FRSUM&$\textbf{83.24}^\dagger$&$\textbf{80.40}^\dagger$&43.54&20.61&40.19\\
     \Xhline{2\arrayrulewidth}
      \rowcolor[gray]{.8}
    T5	&75.32&74.76&\textbf{43.24}&\textbf{20.65}&\textbf{40.18}\\
    ContrastSel&73.55&72.73&43.05&20.45&40.00\\
    CLIFF&74.82&73.31&42.72&19.96&39.41\\
    FRSUM &$\textbf{76.43}^\dagger$&\textbf{74.82}&42.92&20.24&39.69\\
     \Xhline{2\arrayrulewidth}

  \end{tabular}
  
  \caption{Evaluation results of FRSUM on CNN/DM. $\dagger$:  FactCC (\textbf{CC}) or SC \textbf{SC} significantly better than models trained only with NLL (in gray) ($p < 0.05$) in T-test.
 }
  \label{auto_reults_cnn}
\end{table}

\begin{table}[t]
    \small
   \setlength{\tabcolsep}{1.2mm}
   \renewcommand{\arraystretch}{1.3}%调行距
  \begin{tabular}{l|ccccc}
   \Xhline{2\arrayrulewidth}
     &\textbf{CC}&\textbf{SC}& R-1&R-2&R-L  \\
      \Xhline{2\arrayrulewidth}
      \rowcolor[gray]{.8}
      TranS2S&24.15&22.50&30.51&10.30&24.20\\
      Split Encoders&24.78&-&29.45&9.59&23.40\\
      Fact Correction&25.75&-& 36.24&14.37&29.22\\
      FRSUM&$\textbf{28.47}^\dagger$&$\textbf{24.04}^\dagger$&$\textbf{31.38}$&$\textbf{10.89}$&$\textbf{25.01}$\\
      \Xhline{2\arrayrulewidth}
        \rowcolor[gray]{.8}
      BERTSUM&23.81&23.72&38.76&16.33& 31.15\\
      Split Encoders&24.19&-&34.22&13.76&27.86\\
      Fact Correction&\textbf{25.08}&-&36.24&14.37&29.22\\
      FRSUM&24.03&\textbf{23.80}& \textbf{38.79}&\textbf{16.46}&\textbf{31.22}\\
     \Xhline{2\arrayrulewidth}
     \rowcolor[gray]{.8}
     BART&23.64&25.72&\textbf{45.20}&\textbf{21.90}&\textbf{36.88}\\
     ContrastSel&24.89&23.59&44.54&21.23&36.28\\
     CLIFF&23.51&\textbf{25.91}&44.63&21.39&36.43\\
     FRSUM&$\textbf{25.52}^\dagger$&25.80&44.75&21.66&36.76\\
  
%    \rowcolor[gray]{.8}
%    PEGASUS	&23.15&23.33&46.90&	23.57&	38.42 \\
    
%    CLIFF &23.20&21.59&21.59&22.46&37.30   \\
%     ContrastSel&23.24&23.16&46.56&23.17&37.99\\
%    FRSUM & \textbf{23.45}&&\textbf{46.86}&	\textbf{23.68}&	\textbf{38.53}\\
      \rowcolor[gray]{.8}
     \Xhline{2\arrayrulewidth}
    T5	&23.93&22.97&41.15	&	18.18	&	33.10\\
    ContrastSel&26.22&21.58&40.87&17.71&32.68\\
    CLIFF&\textbf{26.72}&22.27&39.48&16.26&31.40\\
    FRSUM &$24.86^\dagger$&\textbf{23.12}&\textbf{41.42} &	\textbf{18.50} & \textbf{33.48} \\
     \Xhline{2\arrayrulewidth}

  \end{tabular}
  
  \caption{Evaluation results of FRSUM on XSum.
 }
  \label{auto_reults_xsum}
\end{table}

\section{Experiment Setup}
\label{exp:info_cnn}

%In this section, we describe the datasets of our experiments and various implementation details.
\subsection{Datasets}
\paragraph{CNN/DM} CNN/DM is a  news dataset with multi-sentence summaries.
CNN/DM contains news articles and associated highlights, which are used as a multi-sentence summary.
We used the standard splits of \citeauthor{hermann2015teaching} for training, validation, and testing (90,266/1,220/1,093 CNN documents and 196,961/12,148/10,397 DailyMail documents). 
We used pre-processed version from \citeauthor{see-etal-2017-get}, and the input documents were truncated to 512 tokens.
\paragraph{XSum}  XSum \cite{narayan-etal-2018-dont} is a news  dataset for extreme summarization,  which
requires the model to summarize a news document with only one sentence summary.
We used the
splits of \citet{narayan-etal-2018-dont} for training, validation, and testing (204,045/11,332/11,334) and followed the pre-processing introduced in their work.
Input documents were truncated to 512 tokens.
 \subsection{Automatic Metric}
\paragraph{Informative Metric}  
We evaluate the informativeness of generated summaries using ROUGE  $F_1$ \cite{lin2004rouge}.
Specifically, we use ROUGE-1 (R-1), ROUGE-2 (R-2) and ROUGE-L (R-L).
\paragraph{Factual Metric}  
We evaluate the faithfulness of the generated summaries by \textbf{FactCC (CC)} \cite{DBLP:conf/emnlp/KryscinskiMXS20}.
Recent large-scale  human evaluation  validates that FactCC correlates  well with human judgments on both  CNN/DM and XSum \cite{pagnoni-etal-2021-understanding}.
We also apply another recent factual metric \textbf{SummaC (SC)}, which achieves the state-of-the-art performance on summary inconsistency detection benchmark \cite{DBLP:journals/corr/abs-2111-09525}.
%ROUGE-L also reasonably correlate with summary faithfulness as judged by human \cite{pagnoni-etal-2021-understanding}.

\subsection{Baselines}\label{baselines}
We select various Seq2Seq models as backbones, based on which we separately  compare FRSUM with different faithful improvement methods.\\
\paragraph{Seq2Seq Models}  We evaluate  FRSUM on extensive baseline systems, especially SOTA pre-trained models.
For non-pretrained models, we select  the vanilla Transformer-based  \cite{vaswani2017attention} Seq2Seq (TranS2S) as the representative.
For pre-trained models, we select the following: partially pre-trained model, BertSum \cite{DBLP:conf/emnlp/LiuL19}; unified  pre-trained model for both language understanding and generation, T5 \cite{raffel2019exploring};  pre-trained model for  language generation tasks, BART \cite{lewis-etal-2020-bart}.%pre-trained model  specifically for summarization, PEGASUS \cite{DBLP:conf/icml/ZhangZSL20}.\\
%We fine-tune these models based on the pre-trained checkpoints.\\
\paragraph{Faithful Improvement}  We also compare against other recent faithfulness improvement methods: \textbf{Split Encoders} \cite{DBLP:conf/aaai/ShahSB20}, a  two-encoder pointer generator;  \textbf{Fact Correction}  \cite{dong-etal-2020-multi-fact}, a QA-based based model that correct the errors in the summary; \textbf{ContrastSel} \cite{DBLP:conf/naacl/ChenZSR21}, a BART-based classifier that  selects faithful summaries in beam search; \textbf{CLIFF} \cite{DBLP:conf/emnlp/Cao021}, a contrastive learning  based training  method utilizing various synthetic augmentation samples.
Please refer to  Appendix \ref{hyper_para} for implementation details.

\begin{table}[t]
    \setlength{\tabcolsep}{0.7mm}
  \renewcommand{\arraystretch}{1.5}
\centering
\small
  \begin{tabular}{l|cccc|cccc}
   \Xhline{2\arrayrulewidth}

    \multirow{2}{*}{Dataset} &
      \multicolumn{4}{c|}{XSum}  &
      \multicolumn{4}{c}{CNN/DM} \\
     &\small{\textbf{CC$\uparrow$}}&\small{\textbf{SC$\uparrow$}} &\small{$E\%\downarrow$}  & \small{R-L} &\small{\textbf{CC$\uparrow$}}&\small{\textbf{SC$\uparrow$}}&\small{$E\%\downarrow$}  & \small{R-L} \\
   \Xhline{2\arrayrulewidth}
    \rowcolor[gray]{.8}
   TranS2S & 24.2&22.5&53.1&24.2 &83.3 &79.1&48.0&35.9\\
   
    +FRSUM &	\textbf{28.5}&\textbf{24.0}&\textbf{49.6}&	\textbf{25.0}&	\textbf{83.8}&\textbf{80.0}&\textbf{43.3}&\textbf{38.0}\\
    \Xhline{2\arrayrulewidth}

    \rowcolor[gray]{.8}
    BertSum & 23.8&24.0&40.1 &31.2&75.7&71.7&33.4&38.8\\

  +FRSUM &\textbf{24.0}&23.8&\textbf{38.5}&\textbf{31.2}&\textbf{77.1}&\textbf{72.2}&\textbf{31.0}&38.7\\
    \Xhline{2\arrayrulewidth}

      \rowcolor[gray]{.8}
    BART	&23.6&25.7&26.7&36.9 & 80.7 &78.7&29.0 & 40.6 \\

    +FRSUM&\textbf{25.5}&\textbf{25.8}&\textbf{24.3} &36.8&\textbf{83.2}&\textbf{80.4}&\textbf{27.5}&40.2 \\
    \Xhline{2\arrayrulewidth}

    %\rowcolor[gray]{.8}
    %PEGASUS	&23.2&22.4&38.4& 79.2 &28.3& \textbf{40.5} \\

    %+FRSUM &
    %\textbf{23.5}&\textbf{20.6}&	\textbf{38.5} &\textbf{79.7}&\textbf{27.8}&40.3\\
      \rowcolor[gray]{.8}
      \Xhline{2\arrayrulewidth}
    T5	&23.9&23.0&37.3	&	33.1	&75.3&74.8&			37.5&40.1\\
    
    +FRSUM &\textbf{24.9} &\textbf{23.1}&\textbf{35.7}& \textbf{33.5} &	\textbf{76.4} &74.8&\textbf{36.4} & 39.7\\
    
    \quad w/o per &24.2&22.8&36.2 &33.2	 &74.0&\textbf{75.0}&37.2& 	40.0\\
   \quad w/o fa & 24.4&22.6&36.3&	33.3&	74.3&72.9&37.0&	39.8\\
   \Xhline{2\arrayrulewidth}

  \end{tabular}
  
  \caption{\textbf{Factual Robustness Analysis} and \textbf{Ablation Study} of FRSUM. $E\%$ denotes the measurements of factual robustness. \textbf{per }and \textbf{fa} refer to factual adversarial permutation and factual adversarial loss.}
  \label{auto_reults_ab}
\end{table}

\section{Results}
Because FRSUM focuses on faithfulness, we expect improvements on factual metrics without harming the performance of informative metrics.
We select the  T5 model for an ablation study and human evaluations because it is a widely used model with a relatively moderate factual robustness $E\%$.  

\subsection{Automatic Evaluation}
\begin{table*}
\begin{subtable}[h]{1\textwidth}
    \centering
  \begin{tabular}{l|c|c|c|c|c|c|c|c}%|c}
  \Xhline{2\arrayrulewidth}
  Model&\textbf{EntE}&\textbf{CircE}&OutE&PredE&OtherE&\#Target &\#Total &Inf\%$\uparrow$\\%&GramE&OtherE&Better Inf.\\ 
  \Xhline{2\arrayrulewidth}
  T5 &24.5&27.0&\textbf{35.0}&16.0&1.0&51.5&103.5&30.0\\%1
  CLIFF&21.0&24.0&38.0&13.0&1.0&45.0&97.0&26.5\\
  ContrastSel&20.0&25.5&40.0&11.0&0.0&45.5&96.5&24.5\\
  FRSUM&\textbf{20.3}&\textbf{22.5}&35.5&14.5&1.5&\textbf{41.0}\small{($18.4\% \downarrow$)}&\textbf{92.5}\small{($10.6\% \downarrow$)}&\textbf{34.0} \\
  \Xhline{2\arrayrulewidth}
  \end{tabular}
  \caption{\footnotesize XSum}
 \end{subtable}

  \begin{subtable}[h]{1\textwidth}
  \centering
   \begin{tabular}{l|c|c|c|c|c|c|c|l}%|c|c}
  \Xhline{2\arrayrulewidth}
  Model&\textbf{EntE}&\textbf{CircE}&OutE&PredE&OtherE&\#Target &\#Total &Inf\%$\uparrow$\\%&GramE&OtherE&Better Inf.\\ 
  \Xhline{2\arrayrulewidth}
  T5 &13.0&14.5&0.5&1.5&0.5&27.5&30.0&40.0\\
  CLIFF&10.0&8.0&1.0&0.5&0.5&18.0&20.0&26.0\\
  ContrastSel&12.0&10&0.5&0.0&0.0&22.0&22.5&22.5\\
  FRSUM&\textbf{6.0}&\textbf{8.0}&2.0&1.5&0.0&\textbf{14.0} \small{($49.1\% \downarrow)$}&\textbf{17.5}\small{($41.7\% \downarrow$)}&\textbf{42.5}\\
  \Xhline{2\arrayrulewidth}
  
  \end{tabular}
 \caption{\footnotesize CNN/DM}
  \end{subtable}
  \caption{Human evaluation results on XSum and CNN/DM, where  the kappa scores are 0.45 and 0.70, respectively. The second to sixth columns report the number of each type of factual errors. \textbf{\#Total} and \textbf{\#Target} report the  number of total types and target types of factual errors, respectively. \textbf{Inf\%} denotes the frequency of the system summary ranked first in informativeness. All the numbers are the average scores of two annotators. 
  Brackets in \#Target and \#Total report the  percentage of relative error decrease of FRSUM over BART.
 }
\label{exp_human}
\end{table*}

The experimental results on CNN/DM and XSum datasets are reported in Table \ref{auto_reults_cnn} and \ref{auto_reults_xsum}.
\textbf{FRSUM} in the last column of each block reports the performance of the baseline further trained with FRSUM.
FRSUM significantly improves the faithfulness of all Seq2Seq baselines and also outperforms all other recent faithfulness improving methods most of the time, achieving the best CC scores on 6 of 8 settings, the best SC scores on 7 of 8 settings. 
Although in some specific cases, some baseline methods such as T5-based CLIFF  and BERTSUM-based Fact Correction on XSUM have higher FactCC scores than FRSUM, their ROUGE scores drop significantly.
By contrast, FRSUM maintains the informativeness of baselines well and even improves the ROUGE scores of several baseline methods, for example, FRSUM improves ROUGE scores of 3 baselines on the XSum dataset.
Overall, compared with other faithfulness improving methods, \emph{FRSUM extensively demonstrates its superiority on improving faithfulness while preserving informativeness}.

\paragraph{Factual Robustness Analysis}
We further analyze the faithfulness of FRSUM considering its effects on  factual robustness  in Table \ref{auto_reults_ab}, where $E\%$ in the table reports the factual robustness of systems.
Concretely, $E\%$ equals to $Mix\%$ in Table \ref{tab:adv}.
According to the results, we can conclude that FRSUM consistently improves the CC score and almost all the SC of all  baseline methods while reducing $E\%$, and thus improves faithfulness.
%For models (TransS2S, BertSumAbs, T5) that are relatively weak at factual robustness ($E\%>30\%$), FRSUM improves their FactCC score over 1 point on both datasets.
%Similarly, for  two models  (PEGASUS and BART) that are relatively robust in factuality ($E\%<30\%$), FRSUM still consistently improves their CC.
\\
\paragraph{Ablation Study}
The results of ablation study are reported in the last two rows in Table \ref{auto_reults_ab}.
\textbf{w/o permut} represents removing the factual adversarial perturbation of FRSUM, and \textbf{w/o fa} represents removing the factual adversarial loss and applying factual adversarial permutations on NLL.
After removing factual adversarial permutations or factual adversarial loss, FRSUM decreases in CC, SC and increases on $E\%$.
Thus, we conclude that these two mechanisms can work separately and combining them further improves the faithfulness.

\subsection{Human Evaluation}
We further conduct human evaluations to assess the effectiveness of FRSUM.  
Instead of comparing systems  pairwise for faithfulness like previous studies, we  report the exact number of different types of factual errors. 
We adopt the linguistically grounded typology of factual errors from Frank \cite{pagnoni-etal-2021-understanding}.
According to Frank, we divide factual errors into 5 types: Entity Error (EntE), Circumstance Error (CircE), Out of Article Error (OutE), Predicate Error (PredE), and Other Error (OtherE).
%In the categorization above,  
EntE and OutE  relate to entity errors, and CircE  mainly relates to numeric errors.
EntE captures entity errors that are contained in the input, while OutE captures entity errors that are not contained in the input.%\footnote{More details can be found in Appendix \ref{error_tpye}. }.
%EntE indicates the generated incorrect entity that contained in the input and  OutE indicates the entity error out of the input. 
For informativeness evaluations, we apply a pairwise comparison between FRSUM  and the original T5.
We invite two professional annotators and randomly select  100  samples from both XSum and CNN/DM test sets for evaluation.

We report the average results  in Table \ref{exp_human}, where ``Inf.'' denotes the ratio of summaries that have a better informativeness than the other systems.
From the number of total errors we can see that FRSUM reduces  factual errors of T5 in both datasets by 10.6\% and 41.7\%, respectively.
 Regarding specific error types, FRSUM substantially reduces EntE and CircE, which are the target types of factual spans
 for  adversarial attacks.
In total, FRSUM reduces the number of target errors by  18.4\% and 49.1\% on XSum and CNN/DM, respectively.
We also notice that models generate a large number of OutEs on XSum.
This is because the XSum dataset itself contains a large number of OutEs in the reference summary while FRSUM is not designed to overcome such noise \cite{DBLP:journals/corr/abs-2102-01672}.

\section{Conclusions and Future Work}
In this paper, we study the faithfulness of abstractive summarization from the new perspective of factual robustness.
We propose a novel adversarial attack method to measure and analyze the factual robustness of current Seq2Seq models.
Furthermore, we propose FRSUM, a faithful improvement training strategy by enhancing the factual robustness of a  model to improves its faithfulness.

\section*{Limitations}
As the first  study on factual robustness, we only analyze entity and number spans which are the most common types of information errors in existing summarization models. 
Future works can study more complicated factual errors such as relation errors.

\section*{Acknowledgments}
We thank the anonymous reviewers for their helpful comments on this paper. This work was partially supported by National Key Research and Development Project
(2020AAA0109703),
National Natural Science Foundation of China (61876009), and National Social Science Foundation Project of China (21\&ZD287).
%Extensive experiments validate the effectiveness of FRSUM in reducing various factual errors.
%FRSUM also demonstrates its potential in further improving and assessing faithfulness of Seq2Seq models with richer adversarial samples.
%In the future work, we will analyze and improve the factual robustness of models on other text generation tasks.

%For future works, we will explore applying factual robustness on
%and demonstrate that it's promising to improve and assess faithfulness of Seq2Seq models by designing richer adversarial samples.

%As the first work to study faithfulness from the perspective of robustness, FRSUM still has great potential to be tapped.
%%%%%%Firstly, we can further study how to construct other factual errors to evaluate and improve models' factual robustness.
%Secondly, future works can explore to effectively defend against adversarial attacks for a more faithful model.
%Finally, factual robustness also has the potential to be applied to faithfulness assessment.
%We expect for  lower $E\%$ and even better adversarial attacks defending method for future research. 
%\section{Conclusion}

\bibliography{anthology,custom}
\bibliographystyle{acl_natbib}
\newpage

\appendix
%\newpage

\section{Hyper-parameter Details}\label{hyper_para}
\begin{table}
\small

     \begin{tabular}{p{0.07\textwidth}>{}p{0.07\textwidth}>{}p{0.07\textwidth}>{}p{0.06\textwidth}>{}p{0.06\textwidth}}
    \Xhline{2\arrayrulewidth}
    Model&Dataset&Training Steps&Learning Rate&Batch Size\\
    \Xhline{2\arrayrulewidth}

    T5&XSum&50k&1e-2&128\\
      &CNN/DM&50k&1e-2&128\\
    \hline
    BART&XSum&20k&5e-5&64\\
        &CNN/DM&15k&5e-5&128\\
    \hline
    PEGASUS&XSum&80k&1e-4&256\\
            &CNN/DM&170k&5e-5&256\\
    \Xhline{2\arrayrulewidth}
    \end{tabular}
    \caption{Parameter settings of pre-train based models used in our experiments}
    \label{tab:para}
\end{table}
\begin{table*}
\small
\centering
    \begin{tabular}{p{0.05\textwidth}>{}p{0.16\textwidth}>{}p{0.35\textwidth}>{}p{0.35\textwidth}}
      \Xhline{2\arrayrulewidth}
         &\textbf{Category}&\textbf{Description}&\textbf{Example}\\ 
          \Xhline{2\arrayrulewidth}
         \textbf{CircE}&Circumstance Error &The additional information (like location or time) specifying the circumstance
around a predicate is wrong.&A \textcolor{red}{22-year-old} teenager has been charged in connection with a serious assault in Bridge Street.\\
         \hline
         \textbf{EntE}&Entity Error &The primary arguments (or their attributes) of the predicate are wrong.&A  teenager has been charged in connection with a serious assault in \textcolor{red}{Aberdeen Sheriff Court}.\\
      \hline
         \textbf{OutE}&Out of Article Error &The statement contains information not
present in the source article.&A teenager has been charged in connection with a serious assault in \textcolor{red}{London}.\\
\hline
\textbf{PredE}&Relation Error &The predicate in the summary statement
is inconsistent with the source article.&A  teenager  \textcolor{red}{is not charged} in connection with a serious assault in Bridge Street.\\
\hline
\textbf{OtherE}&Other Error &Other factual errors like Grammatical Error, Discourse Error.&A  teenager has been charged in \textcolor{red}{connect} with a serious assault in Bridge Street. \textcolor{red}{(GrammarE)}\\
          \Xhline{2\arrayrulewidth}
    \end{tabular}
    \caption{Typology of factual errors in out human evaluation. Original text from the XSum dataset for the examples:\textit{The 22-year - old man needed hospital treatment after the incident on Bridge Street on New Year's Day. Police Scotland said a 15-year - old boy had been charged. The teenager is expected to appear at Aberdeen Sheriff Court.}}
    \label{tab:tpye_fact}
\end{table*}
For TransS2S, we set the number of  both transformer encoder and decoder layers to 6 and the hidden state dimension to 512.
For other pre-training based models, we follow their original parameters for training.  
We apply the base-version  of T5 and large-version of BART and PEGASUS.
The detailed training settings of all the baseline models are set in Table \ref{tab:para}.
We apply beam search for inference. 
During inference, for the XSum dataset, we set beam size to 6, alpha to 0.90, maximum length to 100, maximum length to 10; for CNN/DM dataset, we set beam size to 5, alpha to 0.95, maximum length to 150, maximum length to 30.
For a fair competition, we report the results of CLIFF trained with negative samples constructed by entity swap.

For FRSUM, we apply Spacy for extracting entities and numbers.
In the training process of factual adversarial loss, we randomly sample one $s$ in $y$ for optimization, which we find easier for training.
And we also find a larger size of $A_s$ leads to better performance. 
Thus in practice, we constrain the maximum size of $A_s$ to 10 due to memory constraints.
For time efficiency, we trained the model with  FRSUM  on the checkpoint when the model is close to coverage.
$\eta$ is set to 0.3, $\lambda$ is set to 0.05 and $S$ is the second epoch that the model starts to apply FRSUM for training.
All the models are trained on 8 Nvidia V100 GPUS.

\section{Human Evaluation Details}\label{error_tpye}

\textbf{Typology of Factual Errors} \quad
Recently, \citet{pagnoni-etal-2021-understanding} proposes a typology of factual errors which is theoretically grounded in frame semantics \cite{DBLP:conf/acl/BakerFL98,
DBLP:journals/coling/PalmerKG05},  and linguistic discourse analysis \cite{DBLP:journals/umuai/McRoy00}.
This typology divided factual errors into 7 different categories including  Circumstance Error (CircE), Entity Error (EntE), Out of Article Error (OutE), PredE (Relation Error), Coreference Error (CorefE), Discourse Link Error (LinkE), Grammatical Error (GrammerE).
Because CorefE, LinkE, and GrammerE seldomly appear in generated summaries, in our study, we categorize them jointly as OtherE.
The definitions and examples of typology of factual errors are illustrated in Table \ref{tab:tpye_fact}.\\
\textbf{Annotation Details} \quad Each annotator is first trained to recognize and classify factual errors into a certain category by comparing summaries with the input documents. A summary may contain more than one factual error.
During annotation, each annotator is given a document with two generated summaries from T5 and FRSUM, respectively.
After annotating all the factual errors in these summaries, the annotator also needs to judge which summary is more or equally informative.

\section{Case Study}\label{Human_case}
We show  some cases to demonstrate our human evaluation and the effectiveness of FRSUM in Table \ref{tab:xsum_case} and Table \ref{tab:cnndm_case} on XSum and CNN/DM datasets, respectively.
From Document 1 and Document 2, we illustrate how FRSUM reduces CircE and EntE on XSum.
Document 3 illustrates a special case where the Baseline model generates two errors, OutE and EntE.
Notice that its gold reference also contains OutE, we can infer that the generated OutE  is mainly caused by the  unfaithful reference  in training.
Applying FRSUM on baseline reduces the EntE error but can not reduce the OutE.
Table \ref{tab:cnndm_case} illustrates FRSUM reduces numeric errors (CircE) including date, frequency and score, of  3  examples from  CNN/DM.

\begin{table*}
\small
\centering
    \begin{tabular}{p{0.11\textwidth}p{0.85\textwidth}}

    \Xhline{2\arrayrulewidth}
    \specialrule{0em}{2pt}{2pt}
     \multicolumn{2}{c}{\textbf{XSum Human Evaluation Cases}}\\
     \specialrule{0em}{2pt}{2pt}
     
     \Xhline{2\arrayrulewidth}
     \specialrule{0em}{2pt}{2pt}
     Document 1&\textcolor{red}{The animal had been shot twice in the shoulder and once in its left back leg}, which vets had to amputate.The charity said the one-year-old cat was "incredibly lucky" to survive.Last year the Scottish government held a consultation on licensing air weapons, but a majority of responders opposed the plan.One-year-old Teenie was found injured by her owner Sarah Nisbett in NiddryView, Winchburgh, at about 16:30 on Friday 14 March and taken to the Scottish SPCA.Mrs Nisbett said the cat was now having to learn how to walk again."The gun that was used must have some power because the pellet actually went through her back leg, that's why it was so badly damaged,'' she said."She's now learning how to hop around the house, it's terrible.\textcolor{red}{"The fact that it was three shots is crazy.} We live in a housing estate and there are lots of kids. That just makes it worse because any of them could have been hit in the crossfire."She added: "There's some sick people out there, hopefully somebody will know who's done this and let the police or the Scottish SPCA know."Scottish SPCA Ch Supt Mike Flynn said: "Teenie's owners are understandably very upset and keen for us to find the callous person responsible to ensure no more cats come to harm."This is an alarming incident which only highlights why the Scottish government should implement the licensing of airguns as a matter of urgency."He added: "The new licensing regime should ensure that only those with a lawful reason are allowed to possess such a dangerous weapon. It will also help the police trace anyone using an air gun irresponsibly."...\\
     
     \hline
     \specialrule{0em}{2pt}{2pt}
     Baseline&The Scottish SPCA has appealed for information after a cat was shot \textcolor{red}{twice} in the leg in West Lothian. (\textcolor{red}{CircE})\\
     
    \hline
    \specialrule{0em}{2pt}{2pt}
    +FRSUM&The Scottish SPCA has appealed for information after a cat was shot three times in a crossfire.\\
     \Xhline{2\arrayrulewidth}
    \specialrule{0em}{2pt}{2pt}
    Document 2&It comes in a shake-up of UK military buildings and resettling of regiments.Brecon and Radnorshire Conservative MP Chris Davies condemned the closure, saying there had been a barracks in Brecon since 1805, home to troops who fought the Zulus at Rorke's Drift."This decision is abhorrent and I shall be fighting it every step of the way," he said."The government has a great deal of questions to answer over why it is proposing to close a well-loved and historic barracks in a vitally important military town."Brecon Barracks has served our country with distinction over its long history, with soldiers from the site fighting in every conflict since the early 19th century."This decision shows a blatant lack of respect for that history."Mr Davies said he was launching a petition against the decision, saying the Brecon area had some of the highest unemployment levels in Wales.He also hoped the closure would not damage the town's "thriving" military tourism industry.Brecon barracks has about 85 civilian staff and 90 military but it is not thought jobs are at risk.Mr Davies said he understood the nearby Sennybridge training ground and infantry school at Dering Lines would not be affected.Defence Secretary Sir Michael Fallon told the Commons on Monday the reorganisation in Wales would see a specialist light infantry centre created at St Athan, Vale of Glamorgan. Cawdor Barracks, Pembrokeshire - whose closure was previously announced in 2013 - will now shut in 2024, while a storage depot at Sennybridge will go in 2025.Responding for Labour, Shadow Defence Secretary Nia Griffith, MP for Llanelli, said the ministry was "right to restructure its estate".But she warned closing bases would affect the livelihoods of many people who would face "gnawing uncertainty" over their future.\\
    
     \hline
      \specialrule{0em}{2pt}{2pt}
     Baseline&The government's decision to close \textcolor{red}{military bases in Powys} is " abhorrent ", an MP has said.(\textcolor{red}{EntE})\\
    \hline
     \specialrule{0em}{2pt}{2pt}
     +FRSUM&Plans to close the Brecon Barracks in Powys have been described as " abhorrent ".\\
    \Xhline{2\arrayrulewidth}
     \specialrule{0em}{2pt}{2pt}
     Document 3& Jung won aboard Sam, who was a late replacement when Fischertakinou contracted an infection in July.France's Astier Nicolas took silver and American Phillip Dutton won bronze as GB's William Fox-Pitt finished 12th.Fox-Pitt, 47, was competing just 10 months after being placed in an induced coma following a fall.The three-time Olympic medallist, aboard Chilli Morning, produced a faultless performance in Tuesday's final show-jumping phase.But the former world number one's medal bid had already been ruined by a disappointing performance in the cross-country phase on Monday.He led after the dressage phase, but dropped to 21st after incurring several time penalties in the cross country.Ireland's Jonty Evans finished ninth on Cooley Rorkes Drift.Why not come along, meet and ride Henry the mechanical horse at some of the Official Team GB fan parks during the Rio Olympics?Find out how to get into equestrian with our special guide.Subscribe to the BBC Sport newsletter to get our pick of news, features and video sent to your inbox.\\
    
    \hline
   \specialrule{0em}{2pt}{2pt}
    Gold& \textcolor{red}{Germany's} Michael Jung closed in on a £240,000 bonus prize as he secured a dominant lead to take into the final day of Badminton Horse Trials.    (\textcolor{red}{OutE})\\
    
    \hline
     \specialrule{0em}{2pt}{2pt}
    Baseline & \textcolor{red}{Germany's } Sam Jung won Olympic gold in the equestrian with victory
    \textcolor{red}{ in the dressage phase } on the back of a rider ruled out by illness. (\textcolor{red}{OutE, EntE} )\\
 
    \hline
    \specialrule{0em}{2pt}{2pt}
    +FRSUM & \textcolor{red}{South Africa's } won Olympic gold in the equestrian event at Rio 2016 as Greece's Georgios Fischertakinou was hampered by an infection. (\textcolor{red}{OutE})\\
   
    \Xhline{2\arrayrulewidth}
    
 \end{tabular}
    \caption{Three samples from human evaluations on XSum dataset.}
    \label{tab:xsum_case}
\end{table*}
\begin{table*}
\small
\centering
    \begin{tabular}{p{0.11\textwidth}p{0.85\textwidth}}

    \Xhline{2\arrayrulewidth}
    \specialrule{0em}{2pt}{2pt}
     \multicolumn{2}{c}{\textbf{CNN/DM Human Evaluation Cases}}\\
     \specialrule{0em}{2pt}{2pt}
     
     \Xhline{2\arrayrulewidth}
     \specialrule{0em}{2pt}{2pt}
     Document 4& Lewis Hamilton has conceded to feeling more powerful now than at any stage in his F1 career. It is an ominous warning from a man who has won nine of the last 11 grands prix, been on pole at the last four, and who already holds a 27-point cushion in the drivers' standings. It is no wonder after winning in Bahrain, when Hamilton stepped out of his Mercedes, he immediately stood on top of it and pretended to smack an imaginary baseball out of the circuit. Lewis Hamilton stands on his Mercedes after winning the Bahrain Grand Prix It was another 'home run' performance from Hamilton, a man who claims he is a perfectionist, and who appears to be driving as close to perfection as can possibly be achieved in the sport. It led to the suggestion that perhaps he was feeling unbeatable, to which he replied: 'I don't know what the feeling of being unbeatable is. 'I know I feel very powerful in this car with the package we have, and I feel I'm able to get everything from it. 'I also feel more comfortable in this car than I did in the one last year, and I feel within myself, with whatever approach I have, the power is greater than ever in terms of strength in the car. 'It's like 'yes'. I put the car there and it goes there, I get the exit. 'Naturally you always feel there are things you can still improve on, but it's hard to always pinpoint what those are and to tell the team what to focus on. Hamilton \textcolor{red}{is congratulated by Ferrari rival Kimi Raikkonen on Sunday} 'That's what a team leader has to do, and, along with Nico (Rosberg), I'm trying to guide the team, to tell them where to go next.' ...\\%Last year the Scottish government launched a consultation on licensing air weapons, with a large majority of those who responded opposing the plan.Under the proposed scheme, anyone wanting to own an air gun would need to demonstrate they had a legitimate reason for doing so.A total of 87\% of respondents rejected the idea - with some describing it as "draconian" and "heavy-handed". A small number of people felt ministers were not going far enough.The Scottish SPCA urged anyone with information about the incident to contact them.\\
     
     \hline
     \specialrule{0em}{2pt}{2pt}
     Baseline&"Lewis Hamilton won the Bahrain Grand Prix \textcolor{red}{on Sunday}. The 30-year - old has won nine of the last 11 grands prix. He has been on pole at the last four and holds a 27-point cushion. Hamilton claims he is a perfectionist.
" (\textcolor{red}{CircE})\\
     
    \hline
    \specialrule{0em}{2pt}{2pt}
    +FRSUM&"Lewis Hamilton has conceded to feeling more powerful now than at any stage in his F1 career. The 30-year - old has won nine of the last 11 grands prix and been on pole at the last four. Hamilton holds a 27-point cushion in the drivers'standings.
"\\
     \Xhline{2\arrayrulewidth}
    \specialrule{0em}{2pt}{2pt}
    Document 5& Jermain Defoe says his stunning Wear-Tyne derby winner is justification for his decision to quit MLS and return to the Premier League. The former FC Toronto striker had scored just twice for struggling Sunderland \textcolor{red}{since arriving on £80,000-per-week wages in January.} But Defoe was the hero on Sunday as his 20-yard volley proved enough to secure a fifth straight victory for the Black Cats over Newcastle. Sunderland striker Jermain Defoe believes his stunning volley against Newcastle has proven his worth Defoe's superb first-half strike was enough to secure a 1-0 win for Sunderland in the Wear-Tyne derby Newcastle goalkeeper Tim Krul was completely helpless as Defoe's shot found its way into the top corner The 32-year-old was overcome with emotion in the wake of his brilliant blast, and admits the joy it brought to a sold-out Stadium of Light was too much to take in. ...\\
    
     \hline
      \specialrule{0em}{2pt}{2pt}
     Baseline&Sunderland beat Newcastle 1 - 0 in the Wear - Tyne derby on Sunday. Jermain Defoe scored a stunning first - half volley for the Black Cats. The former FC Toronto striker had scored \textcolor{red}{just twice for the club}.
"(\textcolor{red}{CircE})\\
    \hline
     \specialrule{0em}{2pt}{2pt}
     +FRSUM&"Sunderland beat Newcastle 1 - 0 in the Wear - Tyne derby on Sunday. Jermain Defoe scored a stunning volley in the first half. Defoe had scored just twice for struggling Sunderland since January.
"\\
    \Xhline{2\arrayrulewidth}
     \specialrule{0em}{2pt}{2pt}
     Document 6& Former Valencia striker Aritz Aduriz denied his old team victory with a last-gasp equaliser for Athletic Bilbao at San Mames Stadium. Aduriz pounced in the 90th minute to secure a 1-1 draw after Valencia had been reduced to 10 men. Nicolas Otamendi had harshly received a straight red card eight minutes earlier for a high challenge, and Valencia were unable to hold out in his absence. Athletic Bilbao Aritz Aduriz scored a 90th minute equaliser to deny his former club Valencia victory Substitute Rodrigo De Paul celebrates after coming on to give Valencia the lead Substitute Rodrigo De Paul had given the visitors a deserved lead 16 minutes into the second period, but even though they were unable to claim maximum points, a draw took them above Sevilla into fourth place in their pursuit of possible Champions League qualification. Getafe bolstered their hopes of securing Primera Division status for next season after moving 10 points clear of the bottom three by edging to a 1-0 win at struggling Elche. Fredy Hinestroza scored the only goal of the game with only five minutes remaining in a rare attack for Getafe, after the hosts had spurned a series of chances in the first half. Getafe celebrate after taking a step closer to survival by defeating Elche through Fredy Hinestroza's strike Ecuadorian Felipe Caicedo scored twice for Espanyol in the 3-0 defeat of Villarreal Elsewhere, Espanyol extended Villarreal's winless streak to six games in all competitions as two goals from Felipe Caicedo handed the Catalans a 3-0 win at El Madrigal. Caicedo had a slice of fortune in opening the scoring as Javi Lopez's miscued header deflected off the Ecuadorian and he then thumped home his second of the evening after the break at Sergio Asenjo's near post. Asenjo had a night to forget as he also spilled Sergio Garcia's cross 11 minutes from time to allow Victor Sanchez to roll the ball into an empty net to complete Villarreal's misery.\\

    \hline
     \specialrule{0em}{2pt}{2pt}
    Baseline & "Aritz Aduriz scored a 90th minute equaliser for Athletic Bilbao. The former Valencia striker denied his former club's \textcolor{red}{1-1}  victory. Getafe moved up to fourth in the table after a 1 - 0 win at Elche. Felipe Caicedo scored twice for Espanyol in the 3 - 0 defeat of Villarreal.
" (\textcolor{red}{CircE} )\\
 
    \hline
    \specialrule{0em}{2pt}{2pt}
    +FRSUM & "Aritz Aduriz scored a 90th minute equaliser for Athletic Bilbao. Valencia were reduced to 10 men after Nicolas Otamendi was sent off. Getafe moved 10 points clear of the bottom three with a 1 - 0 win at Elche. Felipe Caicedo scored twice for Espanyol against Villarreal.
"\\
   
    \Xhline{2\arrayrulewidth}
    
 \end{tabular}
    \caption{Three samples from human evaluations on CNN/DM dataset.}
    \label{tab:cnndm_case}
\end{table*}
\end{document}